\documentclass[letterpaper, 10 pt, conference]{ieeeconf}  

\IEEEoverridecommandlockouts                              

\usepackage[utf8]{inputenc} 
\usepackage{graphics} 
\usepackage{epsfig} 
\usepackage{mathptmx} 
\usepackage{times} 
\usepackage{amssymb}  
\usepackage{amsmath} 
\usepackage{algorithm}
\usepackage{algorithmic}
\usepackage{booktabs}
\usepackage{xcolor}
 \usepackage{multirow}

\definecolor{softgreen}{RGB}{0,150,0}
\title{\LARGE \bf
Nav-EE: Navigation-Guided Early Exiting for Efficient Vision-Language Models in Autonomous Driving
}

\begin{document}
\author{Haibo Hu$^{*,1}$, Lianming Huang$^{*,1}$, Xinyu Wang$^{2}$, Yufei Cui$^{2}$,  Shangyu Wu$^{1}$, Nan Guan$^{1}$, Chun Jason Xue$^{3}$
\thanks{*Equal contribution}
\thanks{$^{1}$Department of Computer Science, City University of Hong Kong.
        {\tt\small \{haibohu2-c, lmhuang8-c, shangyuwu2-c\}@my.cityu.edu.hk,}
        {\tt\small nanguan@cityu.edu.hk}}%
\thanks{$^{2}$Department of Computer Science,
        McGill University.
        {\tt\small \{wxy906408362, yufeicui92\}@gmail.com}}%
\thanks{$^{3}$Department of Computer Science,
        Mohamed bin Zayed University of Artificial Intelligence.
        {\tt\small jason.xue@mbzuai.ac.ae}}%
}

\maketitle
\thispagestyle{empty}
\pagestyle{empty}

\begin{abstract}
Vision-Language Models (VLMs) are increasingly applied in autonomous driving for unified perception and reasoning, but high inference latency hinders real-time deployment. Early-exit reduces latency by terminating inference at intermediate layers, yet its task-dependent nature limits generalization across diverse scenarios. We observe that this limitation aligns with autonomous driving: navigation systems can anticipate upcoming contexts (e.g., intersections, traffic lights), indicating which tasks will be required. We propose Nav-EE, a navigation-guided early-exit framework that precomputes task-specific exit layers offline and dynamically applies them online based on navigation priors. Experiments on CODA, Waymo, and BOSCH show that Nav-EE achieves accuracy comparable to full inference while reducing latency by up to 63.9\%. Real-vehicle integration with Autoware Universe further demonstrates reduced inference latency (600\,ms to 300\,ms), supporting faster decision-making in complex scenarios. These results suggest that coupling navigation foresight with early-exit offers a viable path toward efficient deployment of large models in autonomous systems. Code and data are available at our anonymous repository: https://anonymous.4open.science/r/Nav-EE-BBC4
\end{abstract}

\section{INTRODUCTION}

The integration of Vision-Language Models (VLMs) into autonomous driving has emerged as a transformative trend, unifying visual perception with high-level reasoning. Recent advances demonstrate that VLMs can enhance perception and decision-making by capturing contextual cues often missed by conventional models \cite{c1,c2,c3}. For example, Li Auto’s DriveVLM-Dual system provides semantic understanding to assist smaller perception models \cite{c4}, while research prototypes such as Talk2BEV \cite{c1} and Text-to-Drive \cite{c2} highlight the potential to bridge vision and planning. Large-scale models like DeepSeek-VL2 \cite{c5} and LLaVA \cite{c6} further illustrate the strong multimodal reasoning capabilities of modern VLMs. Despite these advances, deploying VLMs in real time remains difficult due to high inference latency and redundant computation, which directly limit responsiveness in safety-critical driving scenarios \cite{c7,c8,c35,c36}.
\begin{figure}[t]
  \centering
  \includegraphics[width=0.5\textwidth]{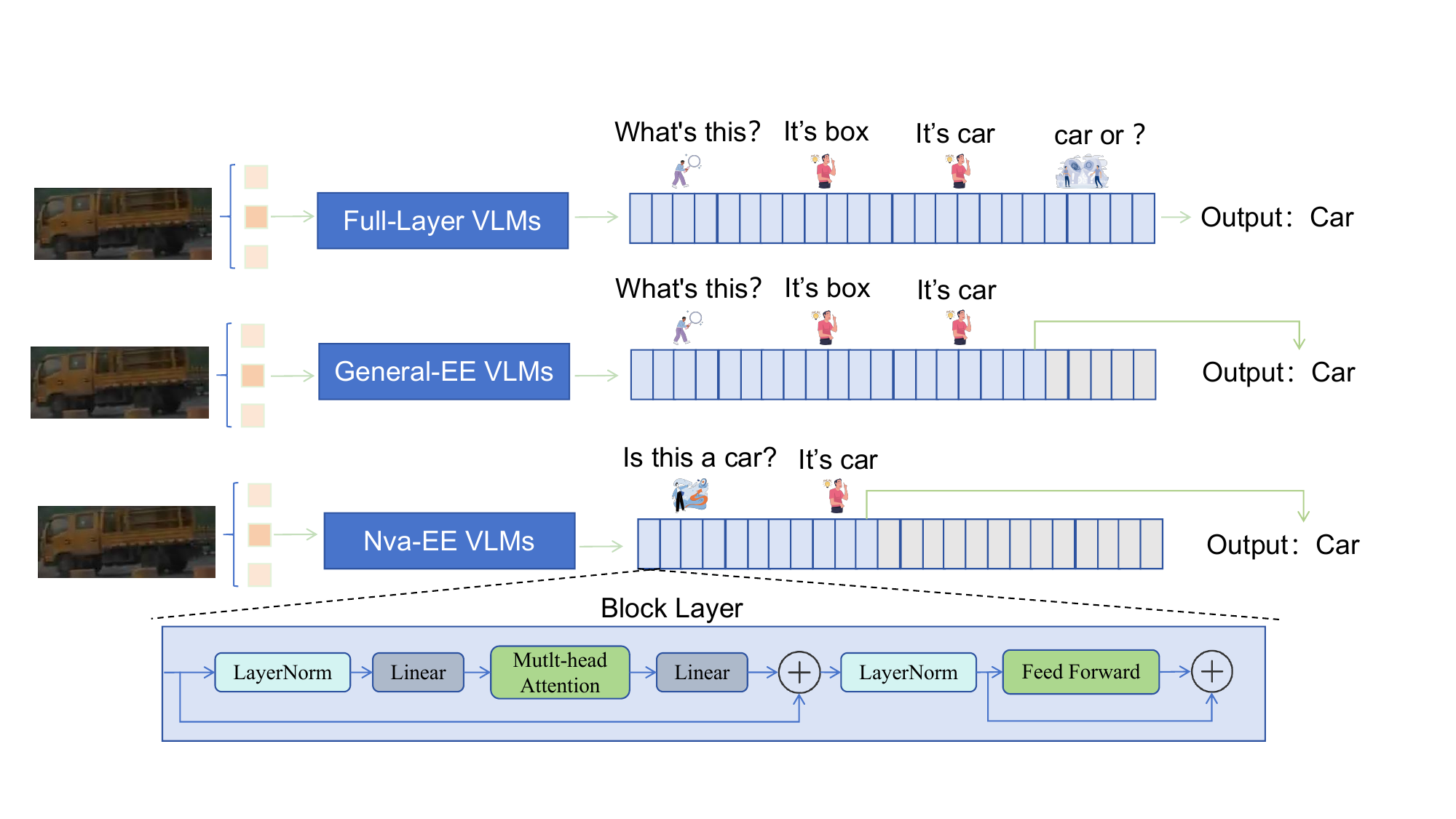}
  \vspace{-20pt}
  \caption{Full-layer inference processes all transformer layers, while generic EE methods stop once predictions stabilize. With navigation priors, VLMs in autonomous driving can adopt more aggressive exits, avoiding redundant computation while preserving accuracy.}
  \vspace{-10pt} 
  \label{pic:vlms_vs}
\end{figure}

\begin{figure}[t]
  \centering
  \includegraphics[width=0.6\textwidth,trim=6cm 0cm 10cm 0cm, clip]{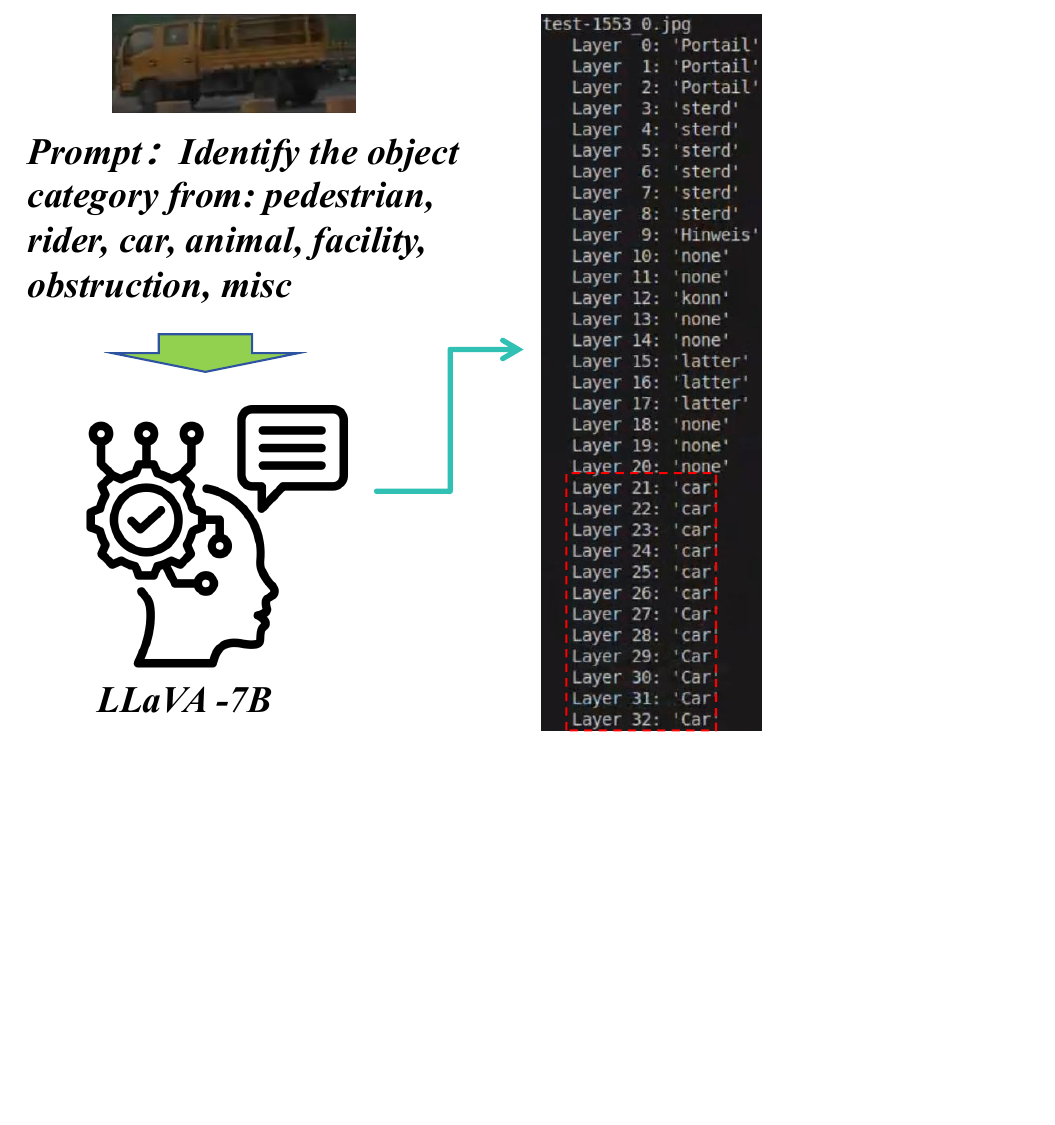}
  \vspace{-130pt}
  \caption{Layer-wise predictions of LLaVA-7B on CODA object recognition, showing early stabilization at the correct label (‘car’).}
  \vspace{-10pt} 
  \label{pic:over_inference}
\end{figure}

Among sparsification techniques, early exit (EE) is particularly attractive, as it halts inference once predictions stabilize \cite{c9,c10,c11}. As shown in Fig.~\ref{pic:over_inference}, correct recognition often emerges in early layers, while later computation adds little value. EE therefore offers strong potential for reducing latency in driving systems, where avoiding unnecessary inference directly improves response time and efficiency.  

However, EE strategies often lack generalization\cite{c37}. Prior work shows that in-domain or task-specific EE can deliver substantial computation savings, whereas out-of-domain or general-purpose benchmarks (e.g., GLUE) see limited benefits \cite{c9,c11,c12}. As illustrated in Fig.~\ref{pic:vlms_vs}, confidence dynamics vary widely across domains, restricting aggressive exits in general-purpose settings. In contrast, for narrow tasks such as text classification or domain-focused applications, EE can skip up to half of the layers with negligible accuracy loss \cite{c9,c12}. This disparity suggests that EE is highly effective in specialized domains but unreliable when applied broadly.  

Interestingly, this limitation aligns naturally with autonomous driving. Navigation systems can anticipate upcoming contexts—such as intersections, traffic lights, or pedestrian zones—thus providing strong priors about the perception or decision-making tasks that will soon be required. As illustrated in Fig.~\ref{pic:nav_show}, these priors enable task-specific EE decisions, allowing VLMs to exit earlier without compromising accuracy.

In this work, we extend EE from unimodal or NLP models to large VLMs for autonomous driving. We exploit the determinism of navigation priors to adopt an aggressive statistical strategy for selecting exit layers, and design a dynamically switchable EE configuration that adapts to driving contexts in real time. We evaluate Nav-EE on large-scale datasets (Waymo \cite{c14}, CODA \cite{c15}, and Bosch) and validate it on a real vehicle integrated with Autoware.Universe. By monitoring HD-map ROS~2 topics to trigger task switches, Nav-EE achieves context-aware early exiting in practice, reducing latency while maintaining or improving accuracy.
\textbf{In summary, the challenge is that conventional EE strategies cannot generalize across the diverse and dynamic tasks in autonomous driving. Our contribution is to adapt EE to large VLMs and leverage deterministic navigation priors as a strength: they decompose driving into predictable sub-tasks, enabling reliable, task-aware acceleration in practice.}

The main contributions of this work are summarized as follows:  

\begin{itemize}
    \item We are the first to adapt early-exit strategies to large VLMs for autonomous driving, transforming them from task-specific techniques into a practical, navigation-aware acceleration mechanism.  
    \item We conduct extensive experiments across diverse datasets and models: the Waymo dataset for common perception tasks, the CODA dataset for animal detection and corner-case recognition, and the Bosch dataset for traffic-light decision making. Results consistently show that Nav-EE reduces inference latency and resource consumption while simultaneously improving accuracy.  
    \item We further validate Nav-EE in real-world deployment by integrating it into an autonomous vehicle running Autoware.Universe. The real-vehicle experiments confirm that navigation-guided early exit can operate reliably under practical driving conditions, achieving both efficiency gains and accuracy improvements.  
\end{itemize}

\section{RELATED WORK}
\subsection{Vision-Language Models in Autonomous Driving}
The integration of Vision-Language Models (VLMs) into autonomous driving has recently gained momentum. 
Unlike conventional perception models, VLMs combine vision and language, enabling richer semantic reasoning about driving scenes. 
Practical systems such as Li Auto’s DriveVLM-Dual \cite{c4}, and research prototypes like Talk2BEV \cite{c1} and Text-to-Drive \cite{c2}, demonstrate how VLMs can support both perception and planning. 
\begin{figure}[H]
  \centering
  \includegraphics[width=0.5\textwidth]{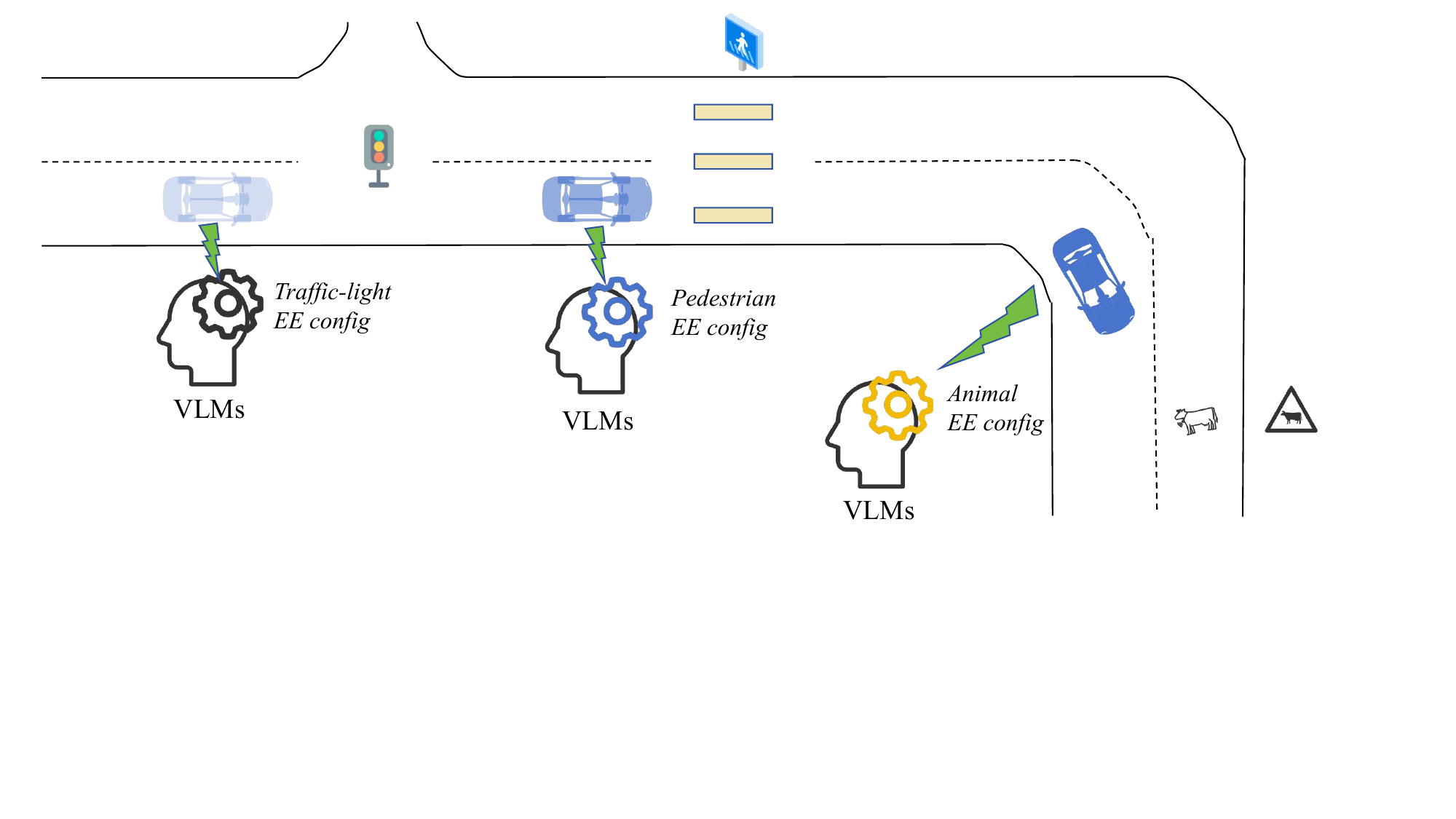}
  \vspace{-60pt}
  \caption{Navigation priors in autonomous driving. HD maps predict upcoming contexts (e.g., traffic-light zones, intersections, pedestrian areas), which Nav-EE uses to trigger task-specific early exits.}
  \label{pic:nav_show}
\end{figure}
Foundation models such as DeepSeek-VL2 \cite{c5} and LLaVA \cite{c6} further highlight the potential of large-scale multimodal reasoning in this domain. Beyond perception, VLMs have been applied to map-based reasoning, natural language navigation, and even traffic law interpretation \cite{c16,c17,c18}. 
These studies suggest that VLMs can provide semantic priors and contextual knowledge that conventional supervised models struggle to capture. 
However, their high computational cost remains a barrier for real-time deployment in safety-critical driving, motivating research into efficiency-oriented techniques such as early exit, pruning, and retrieval augmentation.\cite{c7,c8}.
\begin{figure*}[!t]
\centering
\includegraphics[width=1\textwidth, height=0.4\textheight]{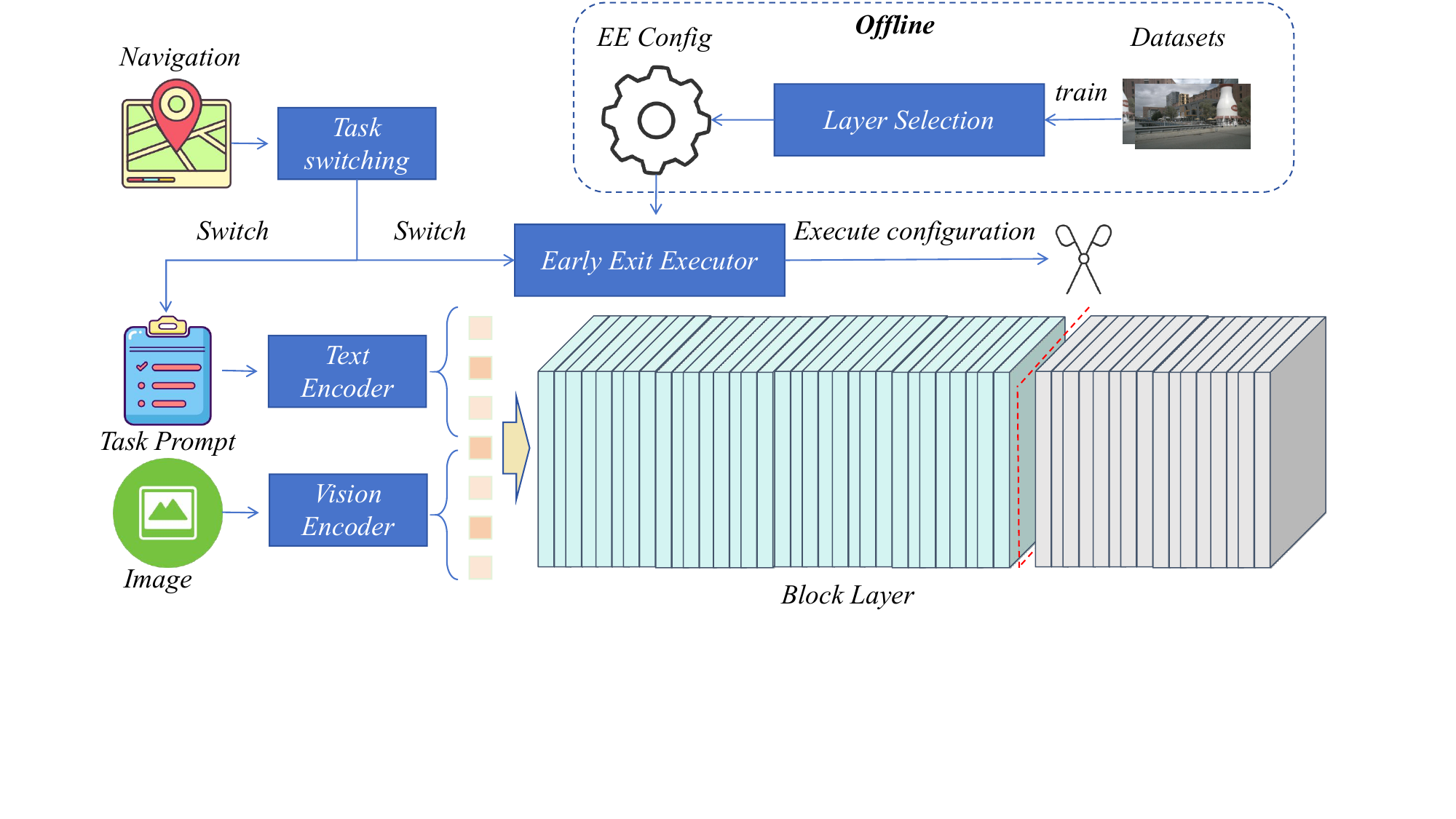}
\vspace{-70pt}
\caption{Overview of Nav-EE: offline profiling identifies task-specific exit layers, which are dynamically triggered by navigation priors for efficient inference in autonomous driving.}
\label{pic:ee_overview}
\end{figure*}
\subsection{Early Exit and Adaptive Inference}
Early exit (EE) reduces inference cost by halting computation once confidence is sufficient, and has been validated across BERT-like backbones and downstream tasks \cite{c9,c10,c11}. Post-deployment acceleration via block bypass and divergence criteria further improves practicality \cite{c12,c13}. For multimodal and generative settings, synchronized/robust EE and confidence-adaptive decoding have also been explored \cite{c18,c17}. Closest to our setting, retrieval-augmented early-exiting (RAEE) shows that instance-wise priors can guide exit-layer selection without re-training the backbone \cite{c16}. Recent work has also explored self-distillation for stabilizing exits \cite{c32}, entropy-based criteria for adaptive decoding \cite{c33}, and layer-wise calibration methods for robust EE under distribution shift \cite{c34}. However, a persistent limitation is that EE often depends on task/domain-specific configurations, hindering generalization. Our work embraces this constraint and turns it into an advantage by leveraging high-certainty navigation priors in autonomous driving to choose aggressive (earlier) exits safely.

\section{METHODOLOGY}
The complete Nav-EE pipeline consists of three stages (Fig.~\ref{pic:ee_overview}).  
First, in the \textbf{offline profiling} stage, we run the VLM across the training dataset and record predictions at every intermediate layer. Based on accuracy statistics, we identify the earliest valid exit layer for each scenario-specific task using an aggressive statistical selection strategy.  
Second, in the \textbf{online inference} stage, the system dynamically switches between these precomputed exit configurations according to navigation-derived scene contexts, ensuring that the most suitable exit layer is activated for each driving condition.  
Finally, in the \textbf{real-vehicle deployment} stage, Nav-EE is integrated into the Autoware.Universe stack, where HD map updates from ROS2 topics trigger real-time task-specific configuration switching.

\subsection{Motivation}
Recent efforts on sparsifying large models for efficiency have explored a wide range of techniques, including parameter pruning \cite{c24}, low-bit quantization \cite{c25} \cite{c30}, and Mixture-of-Experts (MoE) routing \cite{c26}. While pruning and quantization reduce model size or arithmetic cost, they often degrade accuracy or require re-training, and MoE methods introduce additional system-level complexity that is difficult to guarantee in safety-critical driving. In contrast, early-exit offers a simple yet powerful approach: terminating inference at intermediate layers once confident predictions emerge \cite{c9,c10,c11}. Our empirical analysis shows that the most impactful savings arise when exiting entire transformer layers, yielding substantial latency reduction while preserving accuracy. These findings motivate us to adopt early-exit as the core sparsification strategy for autonomous driving.  

\subsection{Problem Formulation}
We consider a Vision-Language Model (VLM) with $L$ transformer layers, denoted as $F=\{f_1,f_2,\dots,f_L\}$. 
For an input $x$, hidden states are computed recursively as $h_l = f_l(h_{l-1})$ with $h_0=\phi(x)$. 
A shared final classification head $g(\cdot)$ is applied to any hidden state, producing the prediction
\[
\hat{y}_l = g(h_l), \quad l=1,\dots,L.
\]
Standard inference computes all layers until $\hat{y}=\hat{y}_L$.

Early-exit (EE) allows inference to terminate at intermediate layers when a confidence criterion is satisfied \cite{c9,c10,c11}:
\[
\Omega(h_l) = 
\begin{cases}
\text{exit at layer } l, & \text{if } M(h_l,T) = 1, \\
\text{continue}, & \text{otherwise},
\end{cases}
\]
where $M(h_l,T)$ is a confidence-matching function relative to threshold $T$.  

Let $\text{Acc}(l)$ and $\text{Lat}(l)$ denote the accuracy and latency when exiting at layer $l$. 
Conventional EE methods adopt conservative thresholds to preserve accuracy, often resulting in late exits and limited efficiency. 
In contrast, we leverage the determinism of autonomous driving tasks to employ an \emph{aggressive statistical strategy}, which selects the earliest valid exit layer by solving:
\begin{equation}
l^\ast = \arg\min_{l} \Big\{ \text{Lat}(l) \;\Big|\; \text{Acc}(l) \geq \text{Acc}(L) \Big\}.
\label{eq:selection}
\end{equation}
This formulation ensures that the selected exit $l^\ast$ preserves full-model accuracy while minimizing latency.

\subsection{Offline Profiling: Aggressive Statistical Layer Selection}
To enable reliable early-exit without retraining, we adopt an \textbf{offline profiling strategy} based on large-scale training data. 
The goal is to evaluate the predictive performance of each intermediate layer and then select the earliest one whose accuracy matches that of the full model. 

Formally, the accuracy at layer $l$ is defined as
\begin{equation}
\text{Acc}(l) = \frac{1}{N} \sum_{i=1}^N \mathbb{1}\!\left[g(h_l(x_i)) = y_i\right],
\label{eq:acc_def}
\end{equation}
where $N$ is the number of training samples, $h_l$ is the hidden representation at layer $l$, $g(\cdot)$ is the shared classification head, and $y_i$ is the ground truth label. 

The complete profiling-and-selection procedure is summarized in Algorithm~\ref{alg:aggressive_selection}.

\begin{algorithm}[b]
\caption{Aggressive Statistical Layer Selection}
\begin{algorithmic}[1]
\REQUIRE Training dataset $\mathcal{D}=\{(x_i,y_i)\}_{i=1}^N$, 
         VLM with $L$ transformer layers $\{f_1,\dots,f_L\}$, 
         shared classification head $g(\cdot)$
\ENSURE Earliest valid early-exit layer $l^\ast$

\STATE \textbf{Initialize:} hidden state $h_0=\phi(x_i)$ for each input $x_i$
\STATE \textbf{Profiling phase:}
\FOR{$l=1$ to $L$}
    \STATE Compute hidden states: $h_l = f_l(h_{l-1})$
    \STATE Generate predictions: $\hat{y}_{l,i} = g(h_l)$ for each $(x_i,y_i)\in \mathcal{D}$
    \STATE Count correct predictions: $c_l \leftarrow \sum_{i=1}^N \mathbb{1}[\hat{y}_{l,i}=y_i]$
    \STATE Estimate accuracy: $\text{Acc}(l) \leftarrow c_l / N$ \COMMENT{as in Eq.~\ref{eq:acc_def}}
\ENDFOR
\STATE Record full-model accuracy: $\text{Acc}(L)$
\STATE \textbf{Selection phase:}
\STATE Identify earliest valid exit according to:
\[
l^\ast = \min \{ l \;|\; \text{Acc}(l) \geq \text{Acc}(L) \}
\]
\RETURN $l^\ast$
\end{algorithmic}
\label{alg:aggressive_selection}
\end{algorithm}
Concretely, each input is passed through the VLM, and predictions are recorded at every intermediate layer $l$. 
We then compute $\text{Acc}(l)$ using Eq.~\ref{eq:acc_def} and compare it against the full-layer accuracy $\text{Acc}(L)$. 
The earliest layer that satisfies $\text{Acc}(l) \geq \text{Acc}(L)$ is selected as the exit point $l^\ast$.  

Suppose a VLM has $L=12$ layers and achieves a full-layer accuracy of 85\% on a validation set. 
When profiling, we obtain the following layer-wise accuracies:
\[
\begin{aligned}
\text{Acc}(6) &= 70\%, \\
\text{Acc}(8) &= 83\%, \\
\text{Acc}(9) &= 85\% \quad (= \text{Acc}(L)), \\
\text{Acc}(10\!-\!12) &\approx 84\% \text{--} 85\%.
\end{aligned}
\]
According to Eq.~\ref{eq:selection}, the earliest valid exit is $l^\ast=9$, 
since it is the first layer that matches the full-model accuracy.  
Thus, inference can safely terminate at layer 9, reducing computation by 25\% without sacrificing accuracy.  
This aggressive statistical strategy enables us to push the exit layer as early as possible, leveraging the deterministic nature of autonomous driving tasks (e.g., vehicle or traffic-light recognition often stabilizes early) to maximize latency reduction without retraining.

\begin{figure}[b]
  \centering
  \vspace{-10pt}
\includegraphics[width=0.5\textwidth]{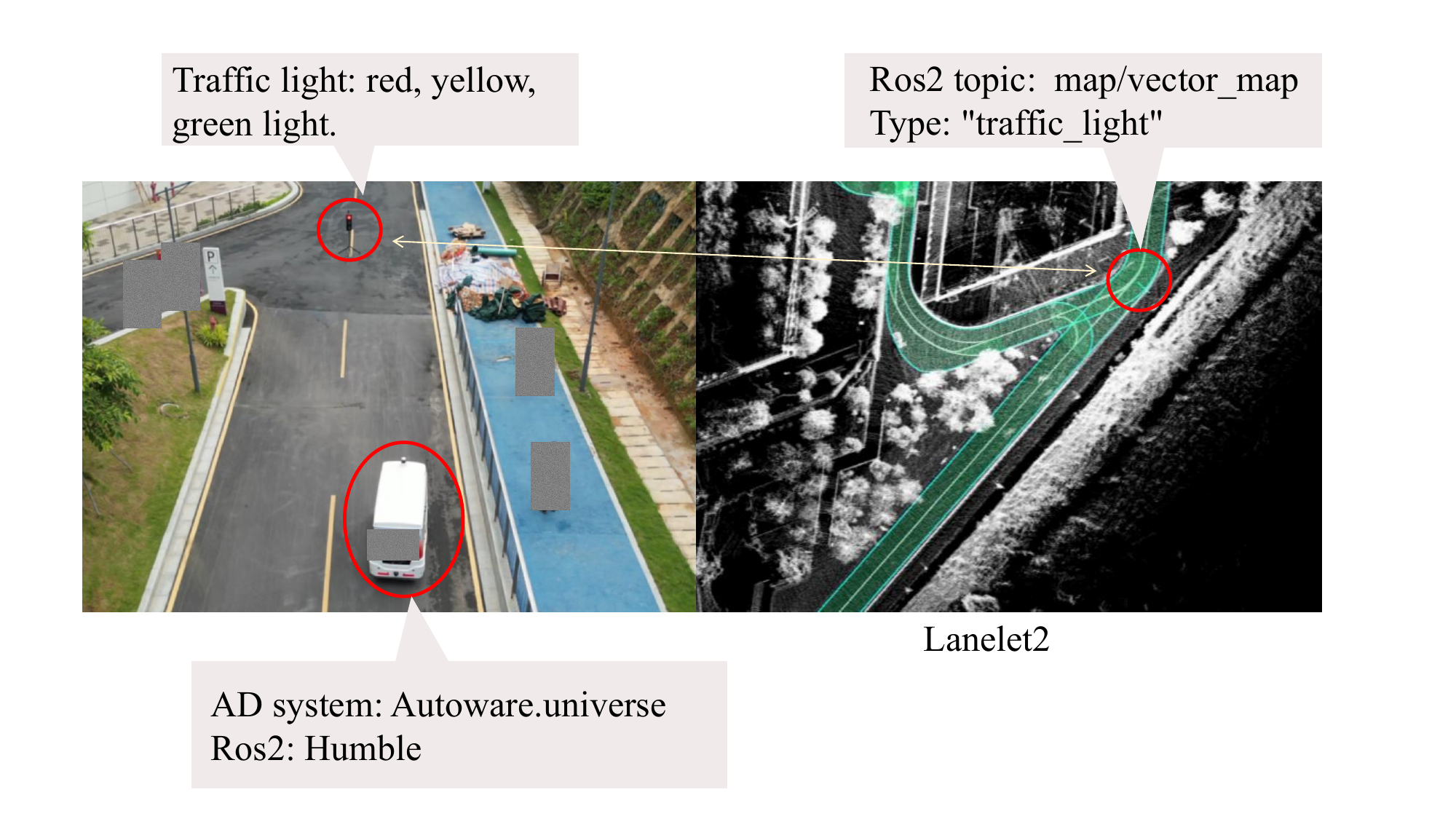}
  \vspace{-20pt}
  \caption{Navigation-aware setup: using Lanelet2 and the ROS2 \texttt{/map/vector\_map} topic to anticipate upcoming traffic lights, enabling VLMs to switch to traffic-light task configurations.}
  \vspace{-10pt} 
  \label{pic:real_overview}
\end{figure}

\subsection{Online Inference: Dynamic Switching for Autonomous Driving}
After the optimal exit layers are profiled offline for different tasks, Nav-EE enables \textbf{online adaptation} through dynamic switching. 
In practice, navigation signals in autonomous driving provide reliable foresight of upcoming contexts (e.g., intersections, traffic lights, or pedestrian zones), as illustrated in Fig.~\ref{pic:nav_show}.
Leveraging this information, the system activates the corresponding pre-computed early-exit configuration for the predicted scene.  

Formally, let $\mathcal{C}=\{c_1,c_2,\dots,c_K\}$ denote the set of scene categories and $\Phi(c_k)$ the offline-determined exit configuration for scenario $c_k$. 
At runtime, when navigation predicts scene $c_k$, inference proceeds under $\Phi(c_k)$, applying the task-specific exit layer $l^\ast_{c_k}$.  

This mechanism ensures that the VLM maintains efficiency and accuracy in diverse contexts. 
Unlike a global configuration that must compromise between tasks, dynamic switching enables scene-aware specialization, making early-exit practical for real-world autonomous driving.

\subsection{Real-Vehicle Deployment: Integration with Autoware.\-Universe}

To validate the practicality of our framework, we integrated Nav-EE into a real autonomous vehicle running the Autoware.Universe platform. The vehicle is equipped with six high-resolution cameras and GPU computing units onboard (RTX 3090), allowing direct deployment of VLM-based perception. 

A key feature of Autoware.Universe is its access to high-definition (HD) maps and navigation signals via ROS2 topics. We subscribe to the HD map topic(/map/vector\_map) to monitor upcoming driving contexts such as intersections, crosswalks, or traffic light regions. These navigation cues are then used to trigger dynamic switching of early-exit configurations: when the HD map indicates a new scene type $c_k$, the system immediately loads the corresponding offline-profiled exit layer $l^\ast_{c_k}$. 
This integration enables Nav-EE to operate seamlessly alongside existing perception and planning modules without requiring modification of the core Autoware stack. As shown in Fig.~\ref{pic:real_overview}
\section{Experiments}
\begin{table}[t]
\vspace{6pt}
\caption{Selected early-exit layers and activated parameters (B). 
For MoE models (DeepSeek-VL2), we report \emph{per-token activated parameters}, 
which are significantly smaller than the total parameter count. 
For dense models (LLaVA-7B), the activated parameters equal the full model size.}
\label{tab:early_result}
\begin{center}
\renewcommand{\arraystretch}{1.3} 
\resizebox{1.0\linewidth}{!}{
\begin{tabular}{llccc}
\toprule
\textbf{Dataset} & \textbf{Tasks} & \textbf{DeepSeek-VL2-Tiny (MoE)} & \textbf{DeepSeek-VL2-Small (MoE)} & \textbf{LLaVA-7B (Dense)} \\
\midrule
 & Full Layers & {\Large 12} ({\small 1.0B}) & {\Large 27} ({\small 2.8B}) & {\Large 32} ({\small 7.0B}) \\
\midrule
\multirow{4}{*}{CODA} 
 & Pedestrian & {\Large 10} ({\small 0.90B}) & {\Large 24} ({\small 2.53B}) & {\Large 18} ({\small 4.07B}) \\
 & Cyclist    & {\Large 10} ({\small 0.90B}) & {\Large 25} ({\small 2.62B}) & {\Large 18} ({\small 4.07B}) \\
 & Vehicle    & {\Large 10} ({\small 0.90B}) & {\Large 20} ({\small 2.18B}) & {\Large 14} ({\small 3.24B}) \\
 & Animal     & {\Large 10} ({\small 0.90B}) & {\Large 21} ({\small 2.27B}) & {\Large 14} ({\small 3.24B}) \\
\midrule
\multirow{2}{*}{WAYMO} 
 & Person     & {\Large 10} ({\small 0.90B}) & {\Large 25} ({\small 2.62B}) & {\Large 17} ({\small 3.86B}) \\
 & Vehicle    & {\Large  9} ({\small 0.85B}) & {\Large 20} ({\small 2.18B}) & {\Large 14} ({\small 3.24B}) \\
\midrule
BOSCH & Traffic-light & {\Large 11} ({\small 0.95B}) & {\Large 22} ({\small 2.36B}) & {\Large 25} ({\small 5.54B}) \\
\bottomrule
\end{tabular}}
\end{center}
\end{table}

\subsection{Experimental Setup}

\subsubsection{Datasets}
We evaluate Nav-EE on three representative autonomous driving datasets.  
\textbf{Waymo Open Dataset} \cite{c14} provides large-scale real-world perception data with common driving scenarios, including vehicles, cyclists, and pedestrians. We adopt it for evaluating standard perception tasks such as vehicle and pedestrian recognition.  
\textbf{CODA Dataset} \cite{c15} focuses on challenging corner cases, covering rare categories such as animals, special vehicles, and unusual obstacles. CODA is particularly valuable for testing the robustness of VLMs under safety-critical conditions and highlights their advantages in recognizing uncommon objects.  
\textbf{Bosch Small Traffic Lights Dataset} is designed for traffic-light state recognition, containing diverse traffic light instances under varying illumination and occlusion conditions. We leverage this dataset to evaluate the decision-critical capability of Nav-EE in traffic signal understanding.  

\begin{table*}[t]
\vspace{6pt}
\caption{Performance of object recognition and decision tasks across datasets. Results are reported for CODA (corner-case perception), WAYMO (regular perception), and BOSCH (traffic-light decision). Improvements are shown relative to full-layer baselines.}
\label{tab:performance_all}
\begin{center}
\resizebox{1.0\linewidth}{!}{
\renewcommand{\arraystretch}{1}
\begin{tabular}{llllllll}
\toprule
\textbf{Models} & \textbf{pedestrian (CODA)} & \textbf{vehicle (CODA)} & \textbf{animal (CODA)} & \textbf{person (WAYMO)} & \textbf{vehicle (WAYMO)} & \textbf{traffic-light (BOSCH)} & \textbf{Avg} \\
\midrule
\multicolumn{8}{l}{\textit{Accuracy (\%)}} \\
LLaVA & 66.93 & 87.28 & 91.18 & 57.28 & 67.33 & 52.10 & 70.35 \\
\textbf{LLaVA\textsubscript{\scriptsize Nav-EE}} & \textbf{100.00} \textcolor{softgreen}{\tiny{(+33.1)}} & \textbf{99.98} \textcolor{softgreen}{\tiny{(+12.7)}} & \textbf{100.00} \textcolor{softgreen}{\tiny{(+8.8)}} & \textbf{98.84} \textcolor{softgreen}{\tiny{(+41.6)}} & \textbf{99.93} \textcolor{softgreen}{\tiny{(+32.6)}} & \textbf{76.93} \textcolor{softgreen}{\tiny{(+24.8)}} & \textbf{95.95} \textcolor{softgreen}{\tiny{(+25.6)}} \\
\cmidrule(lr){1-8}
DS-Tiny & 0.00 & 90.80 & 85.29 & 81.22 & 65.41 & 71.29 & 65.67 \\
\textbf{DS-Tiny\textsubscript{\scriptsize Nav-EE}} & \textbf{34.58} \textcolor{softgreen}{\tiny{(+34.6)}} & \textbf{97.85} \textcolor{softgreen}{\tiny{(+7.7)}} & \textbf{89.71} \textcolor{softgreen}{\tiny{(+4.2)}} & \textbf{94.62} \textcolor{softgreen}{\tiny{(+13.4)}} & \textbf{83.74} \textcolor{softgreen}{\tiny{(+18.3)}} & \textbf{95.80} \textcolor{softgreen}{\tiny{(+24.5)}} & \textbf{82.72} \textcolor{softgreen}{\tiny{(+17.0)}} \\
\cmidrule(lr){1-8}
DS-Small & 0.00 & 98.86 & 77.94 & 56.22 & 98.72 & 52.01 & 63.96 \\
\textbf{DS-Small\textsubscript{\scriptsize Nav-EE}} & \textbf{52.00} \textcolor{softgreen}{\tiny{(+52.0)}} & \textbf{99.44} \textcolor{softgreen}{\tiny{(+0.6)}} & 77.94 & \textbf{68.14} \textcolor{softgreen}{\tiny{(+11.9)}} & \textbf{99.50} \textcolor{softgreen}{\tiny{(+0.8)}} & \textbf{77.55} \textcolor{softgreen}{\tiny{(+25.4)}} & \textbf{79.10} \textcolor{softgreen}{\tiny{(+15.1)}} \\
\midrule
\multicolumn{8}{l}{\textit{Latency (s)}} \\
LLaVA & 0.027 & 0.036 & 0.034 & 0.029 & 0.038 & 0.030 & 0.0323 \\
\textbf{LLaVA\textsubscript{\scriptsize Nav-EE}} & \textbf{0.015} \textcolor{softgreen}{\tiny{($\downarrow$44.4\%)}} & \textbf{0.013} \textcolor{softgreen}{\tiny{($\downarrow$63.9\%)}} & \textbf{0.013} \textcolor{softgreen}{\tiny{($\downarrow$61.8\%)}} & \textbf{0.013} \textcolor{softgreen}{\tiny{($\downarrow$55.2\%)}} & \textbf{0.015} \textcolor{softgreen}{\tiny{($\downarrow$60.5\%)}} & \textbf{0.025} \textcolor{softgreen}{\tiny{($\downarrow$16.7\%)}} & \textbf{0.0157} \textcolor{softgreen}{\tiny{($\downarrow$51.6\%)}} \\
\cmidrule(lr){1-8}
DS-Tiny & 0.096 & 0.093 & 0.095 & 0.100 & 0.094 & 0.098 & 0.0960 \\
\textbf{DS-Tiny\textsubscript{\scriptsize Nav-EE}} & \textbf{0.077} \textcolor{softgreen}{\tiny{($\downarrow$19.8\%)}} & \textbf{0.078} \textcolor{softgreen}{\tiny{($\downarrow$16.1\%)}} & \textbf{0.080} \textcolor{softgreen}{\tiny{($\downarrow$15.8\%)}} & \textbf{0.078} \textcolor{softgreen}{\tiny{($\downarrow$22.0\%)}} & \textbf{0.070} \textcolor{softgreen}{\tiny{($\downarrow$25.5\%)}} & \textbf{0.084} \textcolor{softgreen}{\tiny{($\downarrow$14.3\%)}} & \textbf{0.0778} \textcolor{softgreen}{\tiny{($\downarrow$18.9\%)}} \\
\cmidrule(lr){1-8}
DS-Small & 0.251 & 0.240 & 0.247 & 0.258 & 0.257 & 0.248 & 0.2502 \\
\textbf{DS-Small\textsubscript{\scriptsize Nav-EE}} & \textbf{0.216} \textcolor{softgreen}{\tiny{($\downarrow$13.9\%)}} & \textbf{0.182} \textcolor{softgreen}{\tiny{($\downarrow$24.2\%)}} & \textbf{0.186} \textcolor{softgreen}{\tiny{($\downarrow$24.7\%)}} & \textbf{0.244} \textcolor{softgreen}{\tiny{($\downarrow$5.4\%)}} & \textbf{0.185} \textcolor{softgreen}{\tiny{($\downarrow$28.0\%)}} & \textbf{0.186} \textcolor{softgreen}{\tiny{($\downarrow$25.0\%)}} & \textbf{0.1998} \textcolor{softgreen}{\tiny{($\downarrow$20.1\%)}} \\
\bottomrule
\end{tabular}}
\end{center}
\vspace{-10pt}
\end{table*}

\begin{figure*}[!t]
\centering
\includegraphics[width=1\textwidth, height=0.4\textheight]{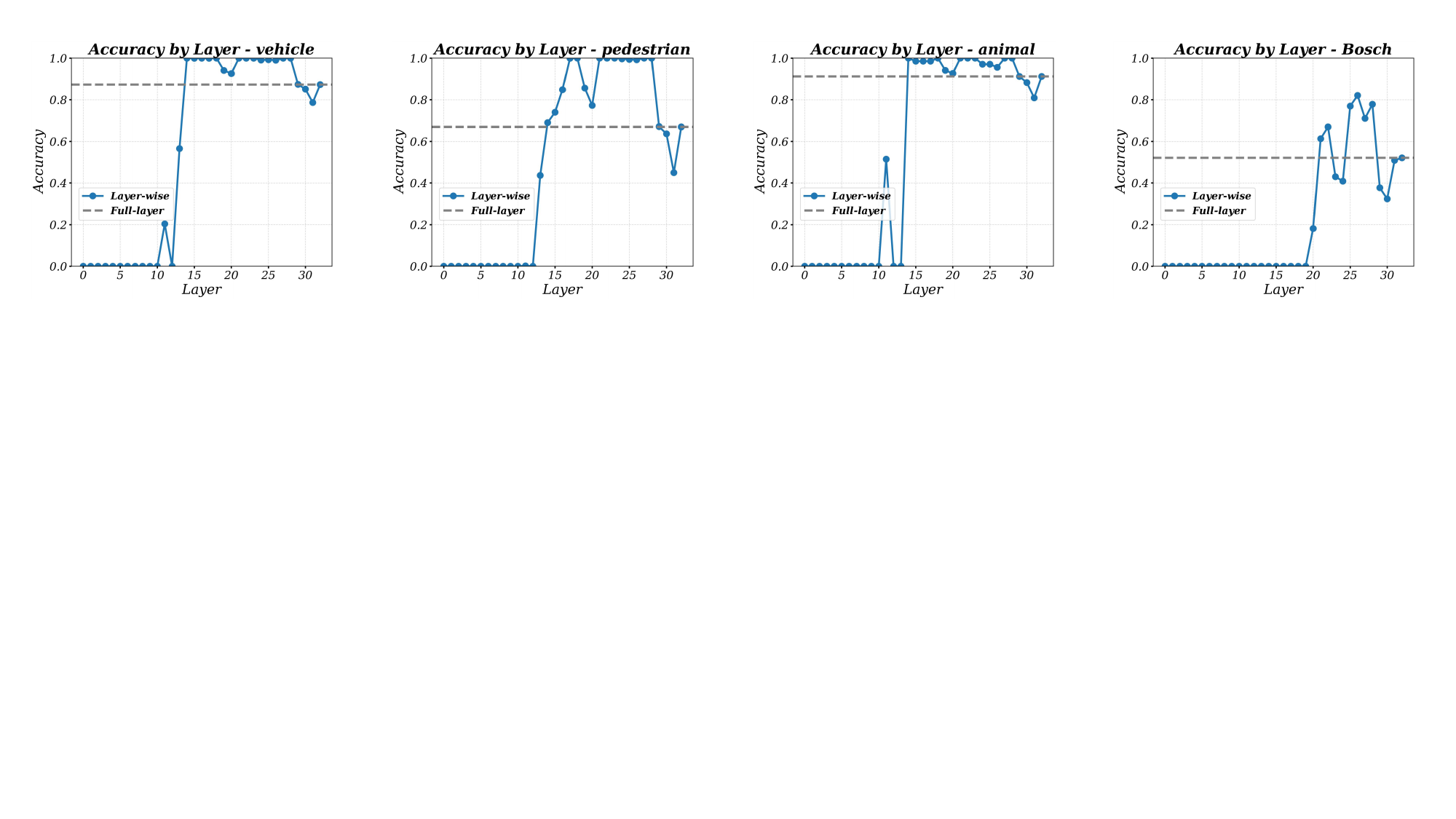}
\vspace{-200pt}
\caption{Layer-wise accuracy dynamics for LLaVA-7B on four representative tasks. The plots for vehicle(CODA), pedestrian(CODA), animal(CODA), and traffic-light (Bosch) recognition show the accuracy at each intermediate layer (solid line) compared to the full-layer baseline (dashed line), illustrating how different tasks stabilize at different model depths.}
\label{pic:ee_finger}
\end{figure*}
\subsubsection{VLMs Configuration}

We adopt three representative Vision-Language Models (VLMs) to evaluate Nav-EE under different model capacities.  
\textbf{LLaVA-v1.6-Vicuna-7B} \cite{c6} is a large-scale VLM with stronger reasoning ability, serving as a testbed for evaluating the scalability of early-exit strategies on high-capacity models.  
Together, these models cover a spectrum from resource-efficient to large-scale VLMs, providing a comprehensive evaluation of Nav-EE across different computational regimes.  
\textbf{DeepSeek-VL2-Tiny} and \textbf{DeepSeek-VL2-Small} \cite{c5} are lightweight variants designed for efficient multimodal reasoning, making them suitable for real-time perception benchmarks.

\subsubsection{Task Selection}
We focus on two types of tasks to evaluate Nav-EE.  
Perception tasks.  
We select sidewalk pedestrian detection, road vehicle detection, road cyclist recognition, and animal recognition.  
On the Waymo Open Dataset, pedestrian and vehicle categories serve as canonical benchmarks for standard perception, reflecting how VLMs perform under typical urban driving conditions.  
On the CODA dataset, however, the same categories (pedestrian and vehicle) are significantly more challenging: as highlighted in the CODA paper \cite{c15}, small perception models often fail to identify these objects under rare, occluded, or unusual conditions. This setting allows us to emphasize the comparative advantage of VLMs and the additional efficiency gains brought by early-exit.  
For animal recognition, CODA provides the most distinctive scenario: animals are rarely included in traditional autonomous driving datasets, making this an out-of-domain task for small perception models. VLMs, in contrast, exhibit stronger generalization to such categories, and our early-exit strategy highlights how efficiency can be preserved even in these rare but safety-critical cases.  Decision task, We additionally evaluate traffic-light state recognition using the Bosch dataset. This task is safety-critical and directly governs driving decisions. Moreover, traffic-light understanding is a canonical case where VLMs can provide semantic guidance to complement smaller perception models, making it a natural testbed for evaluating decision-oriented capabilities.  
By combining both perception and decision tasks, our evaluation spans standard conditions, corner cases, and out-of-domain scenarios, demonstrating that navigation-guided early-exit not only improves efficiency but also amplifies the unique advantages of VLMs in autonomous driving.  

\subsubsection{Metrics}
We report two main metrics:  
1) \textbf{Accuracy}, measured as classification accuracy for corner case objects on CODA and for standard categories on Waymo.  
2) \textbf{Latency}, measured as average inference time per frame in seconds.  
We also provide relative latency reduction compared with full-layer inference.

\subsection{Early Exit Layer Selection Results}
To begin, we analyze the exit layer selection across different perception and decision tasks. 
During the offline profiling stage, each sample from the training dataset is passed through all layers of the VLMs, and the earliest layer that reaches at least $\cdot \text{Acc}(L)$ is recorded as the exit layer. 
Table~\ref{tab:early_result} summarizes the selected layers for major categories including pedestrians, cyclists, animals, vehicles, and traffic lights. 
Lower values indicate earlier exits and therefore higher efficiency. 
Highlighted cells denote particularly aggressive exits where the model can stop significantly earlier without accuracy loss.

We observe that common perception tasks such as vehicle recognition converge at much earlier layers (e.g., 9/12 in Tiny, 14/32 in LLaVA-7B), while more difficult categories like pedestrians and cyclists in CODA require deeper inference. Animal recognition stabilizes at mid-depth, reflecting VLMs’ advantage on out-of-domain tasks. In contrast, traffic-light recognition needs the deepest exits (22–25), consistent with its decision-critical nature. These results highlight that early-exit layers differ systematically across tasks, motivating task-aware and navigation-guided configurations.

In addition to the summary table, Fig.~\ref{pic:ee_finger} further illustrates the \textit{layer-wise accuracy dynamics} for four representative tasks. 
For each case, the solid line reports the accuracy achieved at every intermediate layer, while the dashed horizontal line shows the full-layer accuracy baseline. 
We observe that vehicle recognition quickly stabilizes after about 14 layers, whereas pedestrian recognition requires deeper inference with noticeable fluctuations before convergence. 
Animal recognition reaches stability at mid-depth, suggesting that VLMs can generalize to out-of-domain objects without needing all layers. 
In contrast, traffic-light recognition requires the deepest exits (22--25 layers) before reaching stable accuracy, reflecting its decision-critical nature.

\subsection{Perception Tasks}
We evaluate perception tasks including pedestrians, cyclists, vehicles and animals. 
Table~\ref{tab:performance_all} summarizes both accuracy and latency on the CODA dataset. 
Compared with full-layer inference, our method (Nav-EE) consistently improves accuracy across categories while also reducing inference time. 
This demonstrates that aggressive early-exiting not only avoids redundant computation but also mitigates over-inference, leading to more reliable intermediate-layer representations.

The results on CODA and WAYMO datasets highlight several important trends. 
First, for standard perception tasks such as pedestrian and vehicle recognition in WAYMO, Nav-EE significantly improves both accuracy and latency: DS-Tiny improves from 81.22\% to 94.62\% accuracy with a 22.0\% latency reduction, and LLaVA-7B rises from 67.33\% to 99.93\% accuracy while reducing runtime by more than half. 
Second, for corner-case perception tasks in CODA, early-exit not only maintains but often surpasses full-layer performance. For example, vehicle and animal categories reach nearly perfect accuracy under Nav-EE while saving 15--60\% runtime, showing that VLMs can stabilize predictions earlier without over-inference. 
Third, CODA results further demonstrate that VLMs complement smaller perception models: while DS-Tiny and DS-Small struggle with pedestrians (0\% accuracy in full inference), their performance improves drastically under Nav-EE (e.g., pedestrian detection up to 52\% in DS-Small\textsubscript{\scriptsize Nav-EE}). 
\begin{table}[t]
\vspace{6pt}
\caption{Performance of LLaVA-7B across CODA perception tasks and BOSCH traffic-light decision. We compare full-layer inference, a generic early-exit (General-EE), and our navigation-guided early-exit (Nav-EE). Best results are in bold.}
\label{tab:llava_coda_bosch}
\begin{center}
\renewcommand{\arraystretch}{1.2}
\resizebox{1.0\linewidth}{!}{
\begin{tabular}{lcccc}
\toprule
\textbf{Accuracy (\%)} & \textbf{ped. (CODA)} & \textbf{veh. (CODA)} & \textbf{ani. (CODA)} & \textbf{traff. (BOSCH)} \\
\midrule
LLaVA-7B (Baseline)   & 66.93 & 87.28 & 91.18 & 52.10 \\
LLaVA-7B (General-EE) & 67.11 & 87.38 & 91.18 & 37.68 \\
LLaVA-7B (Nav-EE)     & \textbf{100.00} & \textbf{99.98} & \textbf{100.00} & \textbf{76.93} \\
\midrule
\textbf{Latency (s)}  &       &       &       &       \\
LLaVA-7B (Baseline)   & 0.027 & 0.036 & 0.034 & 0.030 \\
LLaVA-7B (General-EE) & 0.021 & 0.018 & 0.018 & 0.028 \\
LLaVA-7B (Nav-EE)     & \textbf{0.015} & \textbf{0.013} & \textbf{0.013} & \textbf{0.025} \\
\bottomrule
\end{tabular}}
\end{center}
\end{table}

While early-exit is typically motivated by efficiency, our results reveal a critical phenomenon: \textbf{over-thinking in full-layer inference can harm accuracy}. We observe failure cases where mid-depth layers already yield correct predictions, but deeper layers drift toward errors due to noisy refinement or excessive reasoning. This aligns with prior findings in LLMs that deeper inference may induce hallucinations or unstable decision boundaries despite higher cost \cite{c28,c29}. For perception tasks such as vehicles or pedestrians, once high-confidence recognition emerges at intermediate layers, continued inference risks semantic confusion rather than improvement. By terminating at stable early layers, Nav-EE not only improves efficiency but also safeguards against over-inference, delivering faster and more reliable predictions in safety-critical driving scenarios.

\subsection{Traffic-light Decision Results}
Table~\ref{tab:performance_all} reports the performance of traffic-light state recognition on the BOSCH dataset. 
Compared with their full-layer baselines, all models benefit substantially from the proposed Nav-EE strategy. 
For example, DS-Tiny improves from 71.29\% to 95.80\% accuracy while reducing latency from 0.098s to 0.084s. 
Similarly, DS-Small increases accuracy from 52.01\% to 77.55\% with latency reduced from 0.248s to 0.186s. 
The largest improvement is observed with LLaVA-7B, which rises from 52.10\% to 76.93\% accuracy while cutting inference latency from 0.030s to 0.025s.  

These results demonstrate that decision-oriented tasks such as traffic-light recognition, which require semantic reasoning beyond low-level perception, can also benefit from navigation-guided early exits. 
By leveraging task priors, Nav-EE avoids over-computation in later layers while maintaining high reliability, confirming its practical potential for safety-critical decision modules in autonomous driving.  

\subsection{Ablation on General Early-Exit Baseline}
To verify the necessity of navigation-aware configurations, we conduct an ablation study by comparing Nav-EE against a general early-exit (general-EE) baseline. 
The general-EE strategy is implemented by following percentage-based layer reductions that are commonly adopted in LLMs for handling complex tasks, where inference is terminated after a fixed fraction of layers. 
As shown in Table~\ref{tab:llava_coda_bosch}, this baseline only provides marginal improvements over the full-layer inference, and in some cases even reduces accuracy. 
For example, pedestrian recognition improves slightly from 66.93\% to 67.11\%, while traffic-light accuracy decreases from 52.10\% to 37.68\%.  In contrast, our Nav-EE consistently delivers significant gains across all tasks. 
By leveraging navigation priors to guide task-specific exit layers, Nav-EE raises pedestrian, vehicle, and animal recognition to nearly perfect accuracy (100.0\%, 99.98\%, and 100.0\% respectively) with more than 50\% latency reduction. 
For the decision-critical traffic-light recognition task, Nav-EE improves accuracy from 52.10\% to 76.93\% while reducing latency from 0.030s to 0.025s. 
\begin{table}[t]
\vspace{6pt}
\caption{Real-vehicle evaluation with LLaVA-7B. We compare full-layer inference with navigation-guided early-exit (Nav-EE) across tasks. The layer index indicates the exit layer used. Best latencies are in bold.}
\label{tab:llava_real_vehicle}
\begin{center}
\renewcommand{\arraystretch}{1.2}
\resizebox{0.95\linewidth}{!}{
\begin{tabular}{lcccc}
\toprule
\textbf{Method} & \textbf{Person} & \textbf{Traffic-light} & \textbf{Vehicle} & \textbf{Layer Index} \\
\midrule
Baseline & 0.609 & 0.611 & 0.608 & 32 \\
Nav-EE   & \textbf{0.361} & \textbf{0.486} & \textbf{0.295} & 18 / 25 / 14 \\
\bottomrule
\end{tabular}}
\end{center}
\end{table}

These results clearly demonstrate that generic early-exit methods, while useful in language modeling, fail to generalize to multimodal autonomous driving tasks. 
Navigation-aware early exit not only preserves efficiency but also achieves substantial accuracy improvements, validating its effectiveness as a practical deployment strategy.
\subsection{Real-Vehicle Evaluation}
To validate the practicality of Nav-EE in real-world deployment, we integrate LLaVA-7B into an autonomous driving stack running on Autoware.Universe and conduct on-vehicle experiments. 
Table~\ref{tab:llava_real_vehicle} compares the baseline full-layer inference with our Nav-EE configuration across three representative tasks: person recognition, traffic-light recognition, and vehicle recognition. 
The results show consistent latency reductions while maintaining task reliability. 
Specifically, for person detection, Nav-EE reduces the inference time from 0.609s (32 layers) to 0.361s by exiting at layer 18. 
For traffic-light recognition, latency decreases from 0.611s to 0.486s with an exit at layer 25, and for vehicle recognition, Nav-EE exits as early as layer 14, cutting latency from 0.608s to 0.295s.  

These findings demonstrate that Nav-EE can be effectively deployed in real-vehicle settings, offering significant efficiency gains (up to nearly 50\% reduction in latency) without compromising task performance. 
This confirms the feasibility of navigation-guided early exit as a practical mechanism for reducing on-board computation in autonomous driving systems.

\section{Conclusion}
In this work, we introduced \textbf{Nav-EE}, a navigation-guided early-exit framework for Vision-Language Models in autonomous driving. 
By combining offline profiling with online context switching, Nav-EE leverages navigation priors to configure task-specific exit layers, adapting inference to upcoming driving scenarios in a task-aware manner. 
Extensive experiments on CODA, Waymo, and BOSCH show that Nav-EE reduces inference latency while improving recognition accuracy across perception and decision tasks. 
Furthermore, real-vehicle deployment with Autoware.Universe confirmed its practicality, nearly halving end-to-end latency compared to full-layer baselines.  
Overall, Nav-EE demonstrates how predictive navigation can be effectively coupled with early-exit sparsification, enabling resource-efficient and task-aware deployment of large multimodal models under real-time constraints. 
Beyond efficiency, Nav-EE also mitigates the problem of over-thinking in full inference, ensuring more stable and reliable predictions. 
Its design is lightweight and compatible with existing autonomous driving stacks, requiring no retraining of the underlying VLMs. 
Future work will extend this framework toward broader multimodal tasks and integrate it with additional model compression techniques to further enhance scalability.



\begin{thebibliography}{99}
\bibitem{c1} Choudhary T, Dewangan V, Chandhok S, et al. Talk2bev: Language-enhanced bird’s-eye view maps for autonomous driving[C]//2024 IEEE International Conference on Robotics and Automation (ICRA). IEEE, 2024: 16345-16352.  

\bibitem{c2} Nguyen P, Wang T H, Hong Z W, et al. Text-to-drive: Diverse driving behavior synthesis via large language models[C]//2024 IEEE/RSJ International Conference on Intelligent Robots and Systems (IROS). IEEE, 2024: 10495-10502.  

\bibitem{c3} X. Zhou, L. Shan, and X. Gui, “Dynrsl-VLM: Enhancing autonomous driving perception with dynamic resolution vision-language models,” \textit{arXiv preprint arXiv:2503.11265}, 2025. [Online]. Available: https://arxiv.org/abs/2503.11265  

\bibitem{c4} X. Tian, J. Gu, B. Li, Y. Liu, Y. Wang, Z. Zhao, K. Zhan, P. Jia, X. Lang, and H. Zhao, “DriveVLM: The convergence of autonomous driving and large vision-language models,” \textit{arXiv preprint arXiv:2402.12289}, 2024. [Online]. Available: https://arxiv.org/abs/2402.12289  

\bibitem{c5} Z. Wu, X. Chen, Z. Pan, X. Liu, W. Liu, D. Dai, H. Gao, Y. Ma, C. Wu, B. Wang, and others, “DeepSeek-VL2: Mixture-of-experts vision-language models for advanced multimodal understanding,” \textit{arXiv preprint arXiv:2412.10302}, 2024. [Online]. Available: https://arxiv.org/abs/2412.10302  

\bibitem{c6} H. Liu, C. Li, Q. Wu, and Y. J. Lee, “Visual instruction tuning,” \textit{arXiv preprint arXiv:2304.08485}, 2023. [Online]. Available: https://arxiv.org/abs/2304.08485  

\bibitem{c7} Liu X, Zhang M, Chan K H, et al. Low-Latency Edge Learning Framework for Real-Time Decision-Making in Autonomous Driving[J]. Journal of Computer, Signal, and System Research, 2025, 2(3): 75-82.

\bibitem{c8} X. Chen, J. Xu, T. Liang, Z. He, J. Pang, D. Yu, L. Song, Q. Liu, M. Zhou, Z. Zhang, and others, “Do not think that much for 2+3=? On the overthinking of o1-like LLMs,” \textit{arXiv preprint arXiv:2412.21187}, 2024. [Online]. Available: https://arxiv.org/abs/2412.21187  

\bibitem{c9} W. Liu, P. Zhou, Z. Zhao, Z. Wang, H. Deng, and Q. Ju, “FastBERT: a self-distilling BERT with adaptive inference time,” \textit{arXiv preprint arXiv:2004.02178}, 2020. [Online]. Available: https://arxiv.org/abs/2004.02178  

\bibitem{c10} W. Zhu, “LeeBERT: Learned early-exit for BERT with cross-level optimization,” in \textit{Proc. ACL}, 2021, pp. 2968–2980. [Online]. Available: https://aclanthology.org/2021.acl-long.230  

\bibitem{c11} W. Zhou, C. Xu, T. Ge, J. McAuley, K. Xu, and F. Wei, “BERT loses patience: Fast and robust inference with early-exit,” in \textit{Proc. NeurIPS}, vol. 33, 2020, pp. 18330–18341. [Online]. Available: https://proceedings.neurips.cc/paper/2020/hash/
2c063e33f5b1b7a831ed3c9c9a65a949-Abstract.html  

\bibitem{c12} W. Zhu, P. Wang, Y. Ni, G. Xie, and X. Wang, “Badge: speeding up BERT inference after deployment via block-wise bypasses and divergence-based early-exiting,” in \textit{Proc. ACL Industry Track}, 2023, pp. 500–509. [Online]. Available: https://aclanthology.org/2023.acl-industry.54  

\bibitem{c13} T. Sun, X. Liu, W. Zhu, Z. Geng, L. Wu, Y. He, Y. Ni, G. Xie, X. Huang, and X. Qiu, “A simple hash-based early-exiting approach for language understanding and generation,” \textit{arXiv preprint arXiv:2203.01670}, 2022. [Online]. Available: https://arxiv.org/abs/2203.01670  

\bibitem{c14} P. Sun, H. Kretzschmar, X. Dotiwalla, A. Chouard, V. Patnaik, P. Tsui, J. Guo, Y. Zhou, Y. Chai, B. Caine, and others, “Scalability in perception for autonomous driving: Waymo open dataset,” in \textit{Proc. IEEE/CVF Conf. Computer Vision and Pattern Recognition (CVPR)}, 2020, pp. 2446–2454. [Online]. Available: https://arxiv.org/abs/1912.04838  

\bibitem{c15} K. Li, K. Chen, H. Wang, L. Hong, C. Ye, J. Han, Y. Chen, W. Zhang, C. Xu, D.-Y. Yeung, and others, “CODA: A real-world road corner case dataset for object detection in autonomous driving,” in \textit{Proc. European Conf. Computer Vision (ECCV)}, 2022, pp. 406–423. [Online]. Available: https://arxiv.org/abs/2207.05037  

\bibitem{c16}Huang L, Wu S, Cui Y, et al. Raee: A training-free retrieval-augmented early exiting framework for efficient inference[J]. arXiv e-prints, 2024: arXiv: 2405.15198.

\bibitem{c17} T. Schuster, A. Fisch, J. Gupta, M. Dehghani, D. Bahri, V. Tran, Y. Tay, and D. Metzler, “Confident adaptive language modeling,” in \textit{NeurIPS}, 2022, pp. 17456–17472. [Online].

\bibitem{c18} S. Bae, J. Ko, H. Song, and S.-Y. Yun, “Fast and robust early-exiting framework for autoregressive language models with synchronized parallel decoding,” \textit{arXiv preprint arXiv:2310.05424}, 2023. [Online]. Available: https://arxiv.org/abs/2310.05424

\bibitem{c21} A. Feder, K. Keith, E. Manzoor, R. Pryzant, D. Sridhar, Z. Wood-Doughty, J. Eisenstein, J. Grimmer, R. Reichart, M. E. Roberts, and others, “Causal inference in natural language processing: Estimation, prediction, interpretation and beyond,” \textit{TACL}, vol. 10, pp. 1138–1158, 2022. [Online]. Available: https://transacl.org/ojs/index.php/tacl/article/view/4687

\bibitem{c22} J. Pearl, \textit{Causality: Models, Reasoning and Inference}. Cambridge Univ. Press, 2009.

\bibitem{c23} K. Meng, D. Bau, A. Andonian, and Y. Belinkov, “Locating and editing factual associations in GPT,” in \textit{NeurIPS}, 2022, pp. 17359–17372.

\bibitem{c24} S. Han, H. Mao, and W. J. Dally, “Deep compression: Compressing deep neural networks with pruning, trained quantization and Huffman coding,” in \textit{Proc. ICLR}, 2016. [Online]. Available: https://arxiv.org/abs/1510.00149  

\bibitem{c25} Y. Frantar, E. Alistarh, D. Kurtic, M. Denil, and others, “Gptq: Accurate post-training quantization for generative pre-trained transformers,” in \textit{Proc. NeurIPS}, 2022. [Online]. Available: https://arxiv.org/abs/2210.17323  

\bibitem{c26} N. Shazeer, A. Mirhoseini, K. Maziarz, A. Davis, Q. Le, G. Hinton, and J. Dean, “Outrageously large neural networks: The sparsely-gated mixture-of-experts layer,” \textit{arXiv preprint arXiv:1701.06538}, 2017. [Online]. Available: https://arxiv.org/abs/1701.06538  

\bibitem{c27} Ma Y, Cao Y, Sun J, et al. Dolphins: Multimodal language model for driving[C]//European Conference on Computer Vision. Cham: Springer Nature Switzerland, 2024: 403-420. 

\bibitem{c28}Sui Y, Chuang Y N, Wang G, et al. Stop overthinking: A survey on efficient reasoning for large language models[J]. arXiv preprint arXiv:2503.16419, 2025.
\bibitem{c29}Yue L, Du Y, Wang Y, et al. Don't Overthink It: A Survey of Efficient R1-style Large Reasoning Models[J]. arXiv preprint arXiv:2508.02120, 2025.

\bibitem{c30}Wang X, Partovi Nia V, Lu P, Huang J, Chang X W, Chen B, Cui Y. PoTPTQ: A Two-step Power-of-Two Post-training for LLMs[J]. arXiv preprint arXiv:2507.11959, 2025.

\bibitem{c32}Li X, Shao Y, Sun T, et al. Accelerating bert inference for sequence labeling via early-exit[J]. arXiv preprint arXiv:2105.13878, 2021.
\bibitem{c33} Jiang L, Tan Y, Ren J, et al. SAP-BERT: BERT Inference Acceleration Through Skip-Layer and Adaptive Patient Early Exiting[C]//International Conference on Neural Information Processing. Singapore: Springer Nature Singapore, 2024: 269-283.
\bibitem{c34} Mehra A, Seto S, Jaitly N, et al. Understanding the robustness of multi-exit models under common corruptions[J]. arXiv preprint arXiv:2212.01562, 2022.
\bibitem{c35}Huang J, Jin Y, An L, et al. LiteVLM: A Low-Latency Vision-Language Model Inference Pipeline for Resource-Constrained Environments[J]. arXiv preprint arXiv:2506.07416, 2025.
\bibitem{c36}Vasu P K A, Faghri F, Li C L, et al. Fastvlm: Efficient vision encoding for vision language models[C]//Proceedings of the Computer Vision and Pattern Recognition Conference. 2025: 19769-19780.
\bibitem{c37}Zhang Z, Zhu W, Zhang J, et al. PCEE-BERT: Accelerating BERT inference via patient and confident early exiting[C]//Findings of the Association for Computational Linguistics: NAACL 2022. 2022: 327-338.
\end{thebibliography}
\end{document}